\documentclass[10pt,twocolumn,letterpaper]{article}
\usepackage{caption}
\usepackage{subcaption}
\usepackage{iccv}
\usepackage[utf8]{inputenc} 
\usepackage[T1]{fontenc}    
\usepackage[pagebackref=true,breaklinks=true,colorlinks,bookmarks=false]{hyperref}
\usepackage{url}            
\usepackage{booktabs}       
\usepackage{amssymb,amsthm,amsfonts,amsmath}
\usepackage{nicefrac}       
\usepackage{microtype}      
\usepackage{xcolor}         
\usepackage{times}
\usepackage{epsfig}
\usepackage{standalone}
\usepackage{tikz}
\usetikzlibrary{decorations.pathreplacing,calligraphy}
\usetikzlibrary{arrows, calc, positioning,shapes.geometric, arrows.meta,intersections, spath3}
\usepackage{authblk}

\usepackage{multirow}

\newcommand{\mb}[1]{\boldsymbol{#1}}
\newcommand{\mcossim}[0]{S_\text{C}}
\DeclareMathOperator*{\argmax}{arg\,max}

\usepackage{pifont}
\newcommand{\cmark}{\ding{51}}
\newcommand{\xmark}{\ding{55}}

\makeatletter
\DeclareRobustCommand{\rvdots}{%
  \vbox{
    \baselineskip4\p@\lineskiplimit\z@
    \kern-\p@
    \hbox{.}\hbox{.}\hbox{.}
  }}
\makeatother

\iccvfinalcopy 
\def\iccvPaperID{****} 

\ificcvfinal\fi

\begin{document}

\title{Leveraging Vision-Language Foundation Models for Fine-Grained Downstream Tasks}

\author[1]{Denis Coquenet}
\author[1]{Clément Rambour}
\author[1]{Emanuele Dalsasso}
\author[1,2]{Nicolas Thome}
\affil[1]{Conservatoire National des Arts et Métiers, CEDRIC, Paris, France}
\affil[2]{Sorbonne Université, CNRS, ISIR, F-75005 Paris, France}

\maketitle
\ificcvfinal\thispagestyle{empty}\fi

\begin{abstract}
Vision-language foundation models such as CLIP have shown impressive zero-shot performance on many tasks and datasets, especially thanks to their free-text inputs. However, they struggle to handle some downstream tasks, such as fine-grained attribute detection and localization. In this paper, we propose a multitask fine-tuning strategy based on a positive/negative prompt formulation to further leverage the capacities of the vision-language foundation models. Using the CLIP architecture as baseline, we show strong improvements on bird fine-grained attribute detection and localization tasks, while also increasing the classification performance on the CUB200-2011 dataset.
We provide source code for reproducibility purposes: it is available at \url{https://github.com/FactoDeepLearning/MultitaskVLFM}.
\end{abstract}
\section{Introduction}

Foundation models emerged most recently as increasingly large models that leverage high volumes of unlabeled data by being trained with a self-supervised paradigm. The idea is to model as much knowledge as possible with the aim of handling the greatest number of downstream tasks. This desire for genericity goes hand in hand with the need to deal with a variety of modalities (\eg, image and text data), leading to MultiModal Foundation Models (MMFM).

\begin{figure*}[ht!]
    \centering
    \raisebox{0.4in}{\rotatebox{90}{Zero-shot}}
    \begin{subfigure}[b]{0.9\linewidth}
        \centering
        \includegraphics[width=\linewidth]{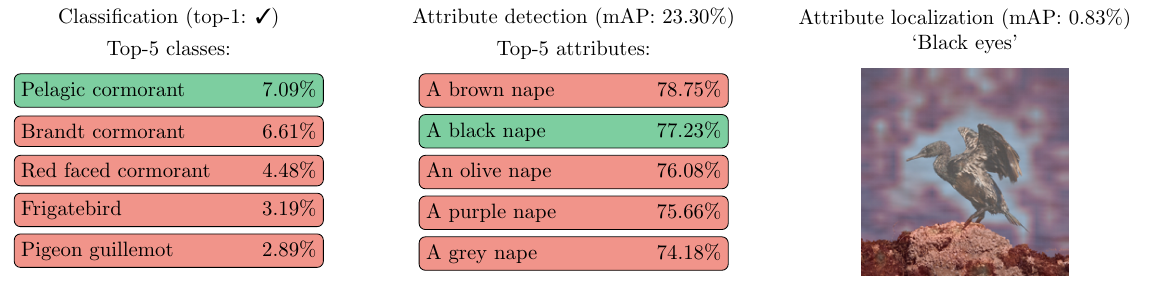}
    \end{subfigure}
    \vspace{0.1cm}
    \hrule
    \rotatebox{90}{Our fine-tuning strategy}
    \begin{subfigure}[b]{0.9\linewidth}
        \centering
        \includegraphics[width=\linewidth]{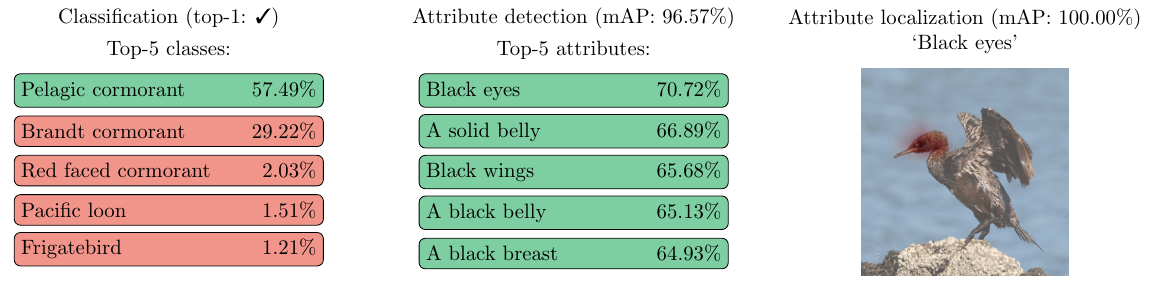}
    \end{subfigure}
    \caption{Comparison between zero-shot and fine-tuned results on an example of the CUB200 dataset. While CLIP generally works well for zero-shot classification, it fails at detecting and localizing fine-grained attributes. The fine-tuning strategy we propose is based on a multitask supervision to leverage CLIP and make it able to deal with these challenging fine-grained tasks. It relies on a positive/negative prompt formulation to improve the generalization on unseen attributes, and we show that MMFM can be used as oracle to reduce the need for localization annotation. Correct and wrong predictions are represented in green and red, respectively.}
    \label{fig:comparison}
\end{figure*}

In this paper, we focus on CLIP \cite{radford2021}, a Vision-Language Foundation Model (VLFM) which aims at aligning image and text modalities in a joint latent space. Despite its impressive zero-shot classification performance on many datasets, CLIP struggles to deal with fine-grained downstream tasks such as attribute detection and localization.
As illustrated in Figure \ref{fig:comparison} (top) for an example from the CUB200 dataset \cite{wah2011}, CLIP succeeds to find the correct bird class in zero shot, but it is not able to detect the presence of fine-grained attributes: it only focuses on the concerned body part and neglect the color, for instance. In addition, since only global representations (whole image with whole text sequence) are brought together through training, CLIP fails to localize attributes precisely in the image.

To tackle these issues, we propose a new multitask fine-tuning strategy to enhance the performance of CLIP for the three tasks: classification, attribute detection and attribute localization. This strategy mainly relies on a hybrid loss to supervise all three tasks, a positive/negative prompt formulation to improve the detection of unseen attributes, and an alignment between text and image patch embeddings. As shown in Figure \ref{fig:comparison} (bottom), this efficient strategy adds coherence to model predictions, by combining natural language explanations, via attribute detection, and visual justification through localization.

In brief, we make the following contributions:
\begin{itemize}
    \item We show that the proposed multitask fine-tuning strategy we propose enables to leverage VLFM for fine-grained downstream tasks, namely classification, attribute detection and attribute localization.
    \item We exploit the concept of positive/negative prompt formulation and demonstrate its ability to improve the detection of attributes unseen during training.
    \item We show that MMFM can be used to alleviate the need for annotations by generating zero-shot pseudo-labels for free, for the localization task.
\end{itemize}

\section{Related works}
\subsection*{Foundation models}

Foundation models first emerged in the Natural Language Processing field with Large Languages Models such as the BERT \cite{devlin2019} family of models, followed by GPT \cite{brown2020}, or LLaMa \cite{touvron2023} most recently. This paradigm mainly relies on self-supervision with masking techniques to overcome the need for large amount of annotated data. The computer vision domain also witnessed the rise of powerful foundation models through MMFM, \eg, text-to-image synthesis with DALL-E \cite{ramesh2021,ramesh2022} or text generation from text/image with Flamingo \cite{alayrac2022}. Models such as Unified-IO \cite{lu2023} and OFA \cite{wang2022} aim at tackling a wide variety of text/image-related tasks by standardizing the way input and output data are formatted and processed. In this paper, we focus on MMFM based on modality alignment, \ie, representations of matching image and text are brought together while representations of unmatching ones are moved away through contrastive learning. CLIP \cite{radford2021} (or ALIGN \cite{jia2021}) models, which belong to this category, offer the opportunity to use any visual or textual input, opening the door for their use for several downstream tasks.

\subsection*{Leveraging CLIP for downstream tasks}
CLIP enables some zero shot applications such as classification or action recognition, as demonstrated in the original paper \cite{radford2021}. However, it is limited by the way it has been trained: it only compares similarities of (text, image) couples. As a result, several works focused on leveraging CLIP capacities to tackle more downstream tasks such as text-guided image generation and synthesis \cite{ramesh2022} or semantic segmentation \cite{zhou2022a,zhou2022b,luddecke2022,ding2022}.
Fine-tuning CLIP for fine-grained attribute recognition was studied in \cite{conde2021} by associating an input prompt by attribute, following the standard contrastive loss. However, CLIP lacks of some kind of humanly-understandable explanation in the decision process.

\subsection*{Explainability}
Interpretability and explainability have been studied from different angles in the literature, notably for reliability purposes. Many works focus on the visual aspect, either through gradient-based approaches \cite{chefer2021,alayrac2022}, or by exploiting the attention weights of Transformer \cite{vaswani2017} attention layers \cite{coquenet2023}. Other works rely on natural language generation to include an explanation part in the output of a visual question-answering model \cite{sammani2022}, to explicitly details the chain of thoughts that lead to a given answer \cite{zhang2023a}, or to list some attributes that lead to the decision \cite{hendricks2016}. Detecting class-related attributes to compute class scores has been studied as another way to add interpretability to the prediction \cite{menon2023,mao2022,yang2022}.
In this paper, classification and attribute detection and localization are performed altogether to enhance the reliability of the prediction.

\section{Modality alignment with foundation models}
Multimodal foundation models such as CLIP \cite{radford2021} and ALIGN \cite{jia2021} focus on projecting the different modalities (namely text and images) into a common feature space, where a given semantic concept share the same representation, no matter the modality. To achieve so, two modules are jointly trained through contrastive learning: an image encoder and a text encoder.

\subsection*{Image encoder}
The image encoder aims at generating a representation of fixed length $d_\text{model}$ from the input image, either by employing a CNN backbone (\eg, ResNet \cite{he2016} or EfficientNet \cite{tan2019}), or a Transformer backbone (\eg, ViT \cite{dosovitskiy2021} or Swin \cite{liu2021}). In this paper, we focus on the latter, as it proved to reach SOTA results \cite{radford2021}.

Let $\mb{X} \in \mathbb{R}^{H \times W \times C}$ be an input image, where $H$ is the height, $W$ the width, and $C$ the number of channels. The image is split into patches of size $P_\text{H} \times P_\text{W}$, and a specific CLS token is added as an additional patch (at index 0), leading to $N+1$ patches with $N=\frac{H}{P_\text{H}} \cdot \frac{W}{P_\text{W}}$.

We denote the image backbone mapping by $f^\mathcal{I}$: $\mb{i} = f^\mathcal{I}(\mb{X})$, where $\mb{i} \in \mathbb{R}^{(N+1) \times d_\text{image}}$ are the representations of the input patches.
An embedding $\mb{e}^\text{i}_X \in \mathbb{R}^{d_\text{model}}$ of the whole image $\mb{X}$ is then generated through projection $g^\mathcal{I}$ of the CLS patch embedding: $\mb{e}^\text{i}_X = g^\mathcal{I}(\mb{i}_0)$.

\subsection*{Text encoder}
Similarly, the text encoder is designed to generate a representation of length $d_\text{model}$, no matter the length of the input text sequence $\mb{u}$.
$\mb{u}$ is first tokenized, then encapsulated with special SOS and EOS tokens, leading to $L$ tokens. Tokens are further embedded through some transformer layers. 

We denote the text backbone mapping by $f^\mathcal{T}$:
$\mb{t} = f^\mathcal{T}(\mb{u})$, with $ \mb{t} \in \mathbb{R}^{L \times d_\text{text}}$.
An embedding $\mb{e}^\text{t}_u \in \mathbb{R}^{d_\text{model}}$ of the whole text sequence is then generated through projection $g^\mathcal{T}$ of the EOS token embedding:
$\mb{e}^\text{t}_u = g^\mathcal{T}(\mb{t}_L)$.

\subsection*{Aligning image and text embeddings\label{sec:aligning}}
Distance between image and text can be computed with the cosine similarity (noted $\mcossim$) between their corresponding embeddings $\mb{e}^\text{i}_X$ and $\mb{e}^\text{t}_u$:
\begin{equation}
    s_u = \mcossim(\mb{e}^\text{i}_X, \mb{e}^\text{t}_u).
\end{equation}

Vision-language foundation models are trained with contrastive loss. Then, they can be applied in zero-shot to, \eg, a classification task, by comparing the similarity score between the embedding of the image $\mb{X}$ at hand and the embeddings of a set of text sequences $\left(\mb{u}_{1},\mb{u}_{2},\dots\right) $, where $\mb{u}_{c}$ refers to a class $c \in \mathcal{C}$.
Probability for each class $c$ can be obtained through softmax activation over the similarity scores $s_c$:
\begin{equation}
    \mb{p}_c = \frac{\displaystyle e^{s_c}}{\displaystyle \sum_{c' \in \mathcal{C}} e^{s_{c'}}}.
\end{equation}

The predicted class $y$ is the one with the highest score:
\begin{equation}
    y = \argmax_c(\mb{p}_c).
\end{equation}

\section{Method}
We propose a multitask fine-tuning approach for CLIP-like foundation models in the context of fine-grained classification. The goal is to enhance the class prediction by justifying it through fine-grained attribute detection and localization. This results in a hybrid loss that gathers the supervision of the three tasks.

\begin{figure*}[ht!]
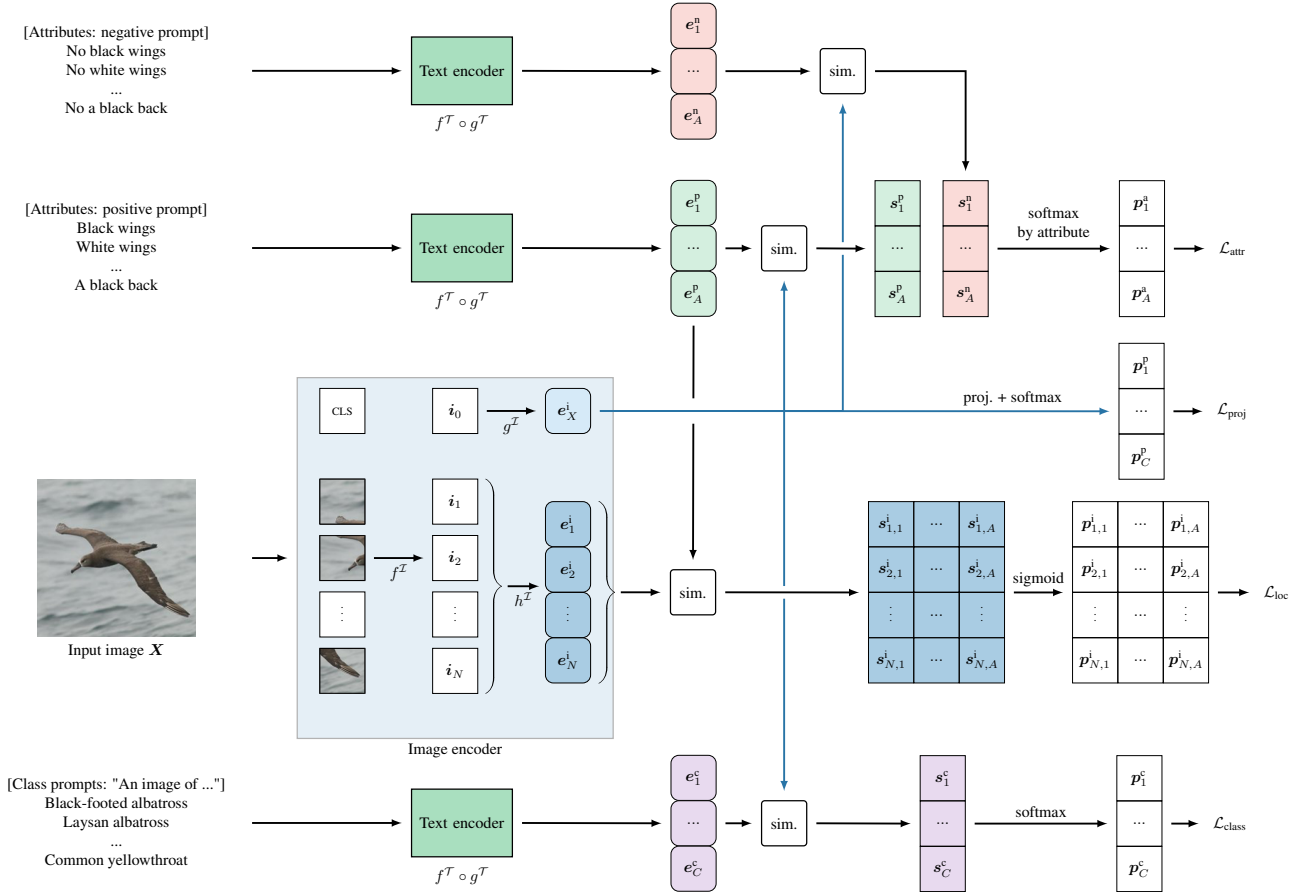

    \centering
    \includestandalone[width=\textwidth]{images/tikz/model_overview}
    \caption{Overview of the architecture and training strategy. Given an input image, the image encoder generates a global image embedding $\mb{e}^\text{i}_X$ and local patch embeddings $\mb{e}^\text{i}_j$. Each class is associated to a dedicated prompt which is further embedded as $\mb{e}^\text{c}_c$ through the text encoder. Similarly, two embeddings are computed for each attribute through the same text encoder: $\mb{e}^\text{p}_a$ for positive formulation and $\mb{e}^\text{n}_a$ for negative formulation. Similarity scores are computed between class and image for classification, between attributes and image for attribute detection, and between patch and positive attributes for localization. An additional projection layer is applied on the image embedding to generate class probabilities, as in traditional classification approaches. The model is trained with a mixture of the four losses induced by these different branches.}
    \label{fig:overview}
\end{figure*}

An overview of the approach is illustrated in Figure \ref{fig:overview}. It is in line with the CLIP paradigm, \ie, we mainly rely on comparison of similarity scores (sim.) between text/image embeddings: image embedding is compared to class prompt embedding for classification, attribute detection is carried out through comparison between image embedding and positive/negative attribute prompt embedding, and patch-level image embeddings are used to localize the attributes. As one can note, the same text encoder (with same weights) is used for all kinds of text input, namely class and attribute prompts. We further details those components in the following. 

\subsection{Classification}
We follow the standard way to carry out classification from CLIP-like models, \ie, we associate a prompt "An image of $c$" for each class $c$ among the class set $\mathcal{C}$.
A text embedding $\mb{e}^\text{c}_c$ is computed through a text encoder for each prompt.
Given an input image $\mb{X}$, a similarity score $\mb{s}^\text{c}_c$ between this image and each class prompt is computed (see section \ref{sec:aligning}). Class probabilities $\mb{p}^\text{c}$ are then obtained through softmax activation.
For this task, we use the Cross-Entropy (CE) loss over the class probabilities, leading to the dedicated loss $\mathcal{L}_\text{class} = \mathcal{L}_\text{CE}(\mb{p}^\text{c}, \mb{c}^*) = -\sum_{c \in \mathcal{C}} \mb{c}^*_c \log{\mb{p}^\text{c}_c}$, where $\mb{c}^*$ is the one-hot encoded ground truth class.

We also integrate an additional projection layer, applied on the image embedding, in order to compute the class probabilities $\mb{p}^\text{p}$ and improve the classification accuracy. It implies the use of the second loss: $\mathcal{L}_\text{proj}^\text{c} = \mathcal{L}_\text{CE}(\mb{p}^\text{p}, c^*)$. The contribution of this component is shown empirically in Section \ref{section:ablation-study}.

\subsection{Attribute detection}
\label{section:method-attributes}
Attributes can be defined as visual features, formulated through natural language, that match a class and a specific image: it can correspond to colors, shapes, or components. 
Some attributes can match a class without being in a corresponding image due to angle or cropping, for instance. The attribute set is noted $\mathcal{A}$. Based on \cite{pellegrini2023}, we opted for a positive/negative prompt formulation of the attribute prompts, \ie, each attribute is associated to two prompts: a positive prompt which corresponds to the attribute itself, and a negative prompt whose keyword \textit{no} is added before the attribute. For both positive and negative attributes, we compute similarity scores ($\mb{s}^\text{p}_a$ and $\mb{s}^\text{n}_a$) with respect to the input image. Attribute probabilities $\mb{p}^\text{a}$ are obtained by applying a softmax activation on each couple of positive/negative score. This attribute detection part is trained through the loss $\mathcal{L}_\text{attr} = \frac{1}{|\mathcal{A}|} \sum_{a \in \mathcal{A}} \mathcal{L}_\text{CE}(\mb{p}^\text{a}_a, \alpha_a)$, where $\alpha_a$ is the ground truth associated to attribute $a$.

\subsection{Localization}

To enhance the reliability of the attribute detection, we propose to localize them in the input image. Segmentation annotation being costly to produce, we focus on coarse localization at patch level, \ie, selecting the image patches that match a given attribute. 
To this end, we take advantage of the vision-language architecture by comparing image patch embeddings $\mb{e}^\text{i}_j$ with an attribute embedding $\mb{e}^p$ (in positive formulation). 

However, in CLIP-like architectures, the projection layer $g^\mathcal{I}$ is only used on the CLS token embedding; this implies that the patch embeddings $\mb{e}^\text{i}_j \in \mathbb{R}^{d_\text{image}}$ and the attribute embedding $\mb{e}^p \in \mathbb{R}^{d_\text{model}}$ do not share the same representation space. To overcome this issue, we introduce a new projection layer $h^\mathcal{I}$ to project the patch embeddings into the same feature space as the attribute embedding: $\mb{e}^\text{i}_j = h^\mathcal{I}(\mb{i}_j)$, with $\mb{e}^\text{i}_j \in \mathbb{R}^{d_\text{model}}$.
Similarity scores $\mb{s}^\text{i}_{j,a}$ are computed through cosine similarity for each patch $j$ and attribute $a$. The associated probabilities $\mb{p}^\text{i}_{j,a}$ can be obtained through sigmoid activation.

The patch-level localization is trained using the Binary Cross-Entropy loss ($\mathcal{L}_\text{BCE}$) over the different patches and for all the attributes present in the input image:
\begin{equation}
\mathcal{L}_\text{loc} = \frac{1}{|\mathcal{A}|} \sum_{a \in \mathcal{A}} \alpha_a \cdot \left( \frac{1}{N} \sum_{j=1}^N \mathcal{L}_\text{BCE}(\mb{p}^\text{i}_{j,a}, \mb{l}^*_{j,a}) \right),
\end{equation}
where $\mb{l}^*_{j,a}$ is the binary ground truth location for patch $j$ and attribute $a$, and $N$ is the number of patches.

\begin{table*}[ht!]
    \caption{Performance comparison between Zero-Shot (ZS) and the multitask Fine-Tuning (FT) strategy we propose on the CUB200 dataset.}
    \label{tab:zeroshot}
    \centering
        \resizebox{\linewidth}{!}{
        \begin{tabular}{c c c c c c}
            \hline
            Backbone & Pre-training dataset & \multirow{2}{*}{Approach} & Top-1 & Attribute  & Attribute \\
            (\# params.) & (\# samples) & & accuracy (\%) & detection mAP (\%) & localization mAP (\%)\\
            \hline
            \hline
            CLIP ViT-L/14 & WIT & ZS & 62.65 & 22.69 & 9.22\\
            (429 M) & (400 M) & FT & 86.40 & \textbf{68.49} & 67.57\\
            \hline
            CLIP ViT-L/14 & LAION-2B & ZS & 75.15 & 22.39 & 8.73\\
            (429 M) & (2.32 B) & FT & 86.28 & 66.11 & 61.73\\
            \hline
            CLIP ViT-H/14 & LAION-2B & ZS & 81.98 & 19.98 & 7.69\\
            (988 M) & (2.32 B) & FT & 87.99  & 67.42 & 65.93\\
            \hline
            \hline
            Swin-L \& CLIP text & ImageNet-21K \& LAION-2B & ZS & 0.17 & 15.34 & 5.81 \\
            (196 M \& 125 M) & (144 M \& 2.32 B) & FT & \textbf{91.20} & 63.78 & \textbf{74.85} \\
            \hline
        \end{tabular}
        }
\end{table*}

\subsection{Hybrid loss}
The three tasks (classification, attribute detection and localization) are trained altogether in an end-to-end fashion through a hybrid loss:
\begin{equation}
    \mathcal{L} = \lambda_\text{class} \mathcal{L}_\text{class} + \lambda_\text{proj} \mathcal{L}_\text{proj} + \lambda_\text{attr} \mathcal{L}_\text{attr} + \lambda_\text{loc} \mathcal{L}_\text{loc},
\end{equation}
where $\lambda_\text{class}$, $\lambda_\text{proj}$, $\lambda_\text{attr}$, and $\lambda_\text{proj}$ are set to 1 if not stated otherwise. The choice of the loss weights is discussed in Section \ref{section:ablation-study}.
\section{Experiments}

\subsection*{Dataset}
Experiments are conducted on CUB200-2011 fine-grained bird classification dataset \cite{wah2011}. It includes 5,994 training images and 5,794 testing examples. Each image is associated to one of the 200 bird classes. The body parts of the birds are localized through 2D point annotations referring to their center. Given that image encoders subsample the input images into patches, we use these annotations to define a patch-level ground truth to train the localization part. In addition, 312 binary attributes are annotated per image, mainly as body-part colors, patterns, and shapes. An annotator confidence level is associated to each attribute annotation (from 1 to 4): we only keep annotations associated to a confidence level of 3 (probably) or 4 (definitely).

\subsection*{Implementation details}
Models are trained for 100 epochs on a single GPU V100 (32Gb) with gradient accumulation over 200 samples between each back-propagation. As for the original CLIP pre-training \cite{radford2021}, we use the Adam optimizer with $\beta_1=0.9$, $\beta_2=0.98$, and $\epsilon=10^{-6}$, as well as a clamped temperature factor (applied on cosine similarity scores) with an upper limit of 100. The initial learning rate is set to $10^{-5}$, and we use automatic mixed-precision. 
We preserve the pre-processing steps of the pre-training stage, \ie, images are resized through bi-linear interpolation and center cropping to a size of $224 \times 224$ for ViT backbones and $384 \times 384$ for Swin ones \cite{liu2021}. Data augmentation is applied during training to input images using random cropping and horizontal flip.

\subsection*{Metrics}
We use the top-1 accuracy to evaluate the classification performance. The attribute detection performance is computed through a mean Average Precision (mAP) over all the 312 binary attributes, given that several attributes can match a given image. In the same way, the attribute localization is evaluated through mAP over the image patches, for each correct attribute.

\subsection{Leveraging foundation models through multitask fine-tuning}

In this first experiment, we compare the performance of CLIP models for the three tasks (classification, attribute detection and localization) with different backbones and pre-training datasets. Table \ref{tab:zeroshot} shows the results in zero-shot configuration compared to the multitask fine-tuning strategy that we propose. Fine-tuning is carried out on the target dataset (CUB200) and only the parameters of the last layers of both image and text encoders (\ie, from the last transformer layer) are trained, for training stability.
As one can note, classification accuracy increases with the size of both the backbone and the pre-training dataset. While the classification gap between zero-shot and fine-tuning tends to close up, the opposite applies for the attribute detection: zero-shot models fail to detect fine-grained attributes in the image. The gain is substantial with fine-tuning (from 19.98\% to 67.42\% of mAP for the larger ViT-H CLIP model). 
Standard CLIP architecture does not enable to compare attribute and image patch embeddings due to dimensionality mismatching. This way, PCA is used in the zero-shot configuration to project embeddings onto a space of the same dimension (30 here, obtained through elbow method applied on the eigenvalues). As one can notice, introducing a single projection layer for the image patch embeddings while fine-tuning enables to increase the attribute localization mAP by far more than 53 points in any case.

\begin{figure*}[ht!]
    \centering
    \raisebox{0.4in}{\rotatebox{90}{Black eyes}}
    \begin{subfigure}[b]{0.19\linewidth}
        \centering
        \includegraphics[width=\linewidth]{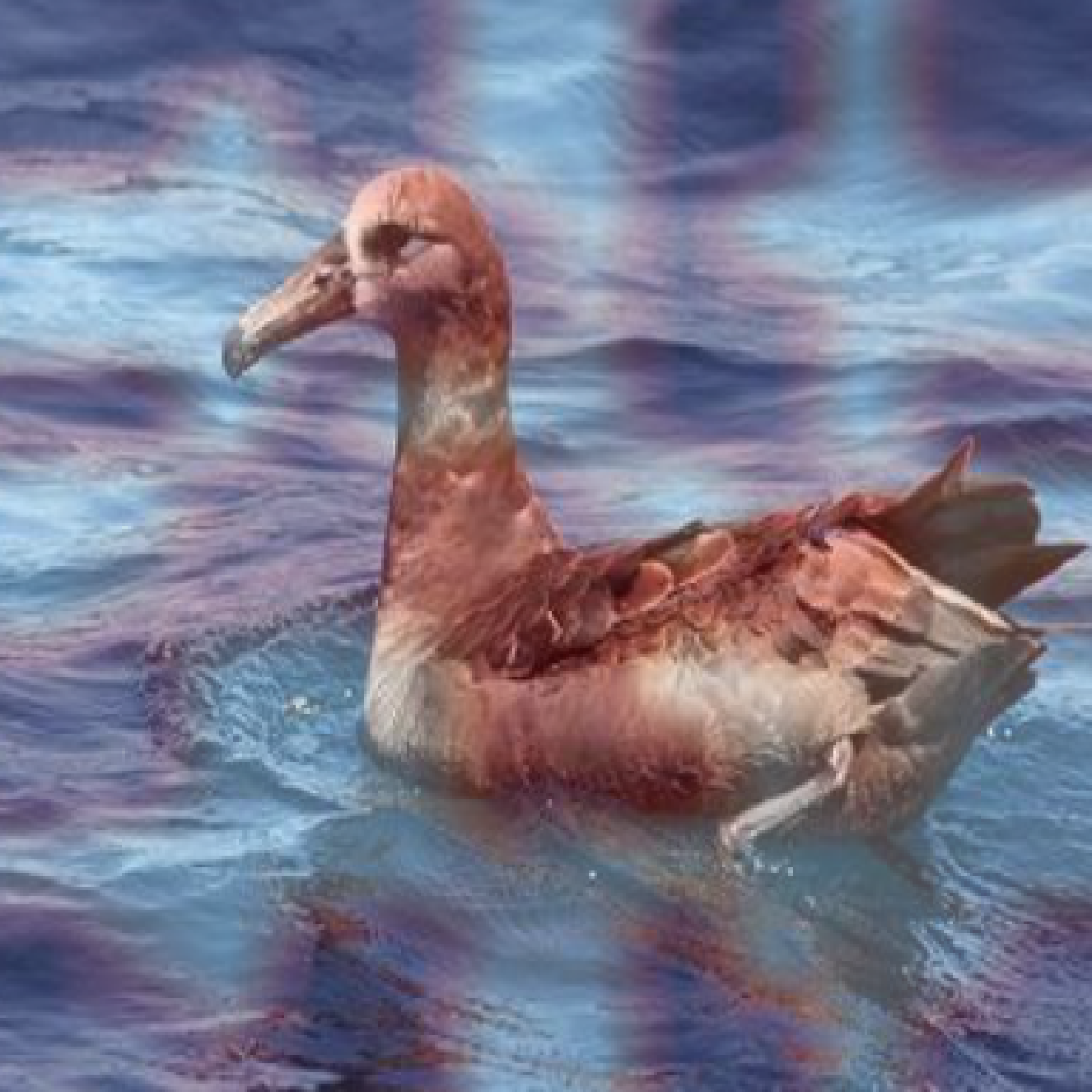}
    \end{subfigure}
    \hfill
    \begin{subfigure}[b]{0.19\linewidth}
        \centering
        \includegraphics[width=\linewidth]{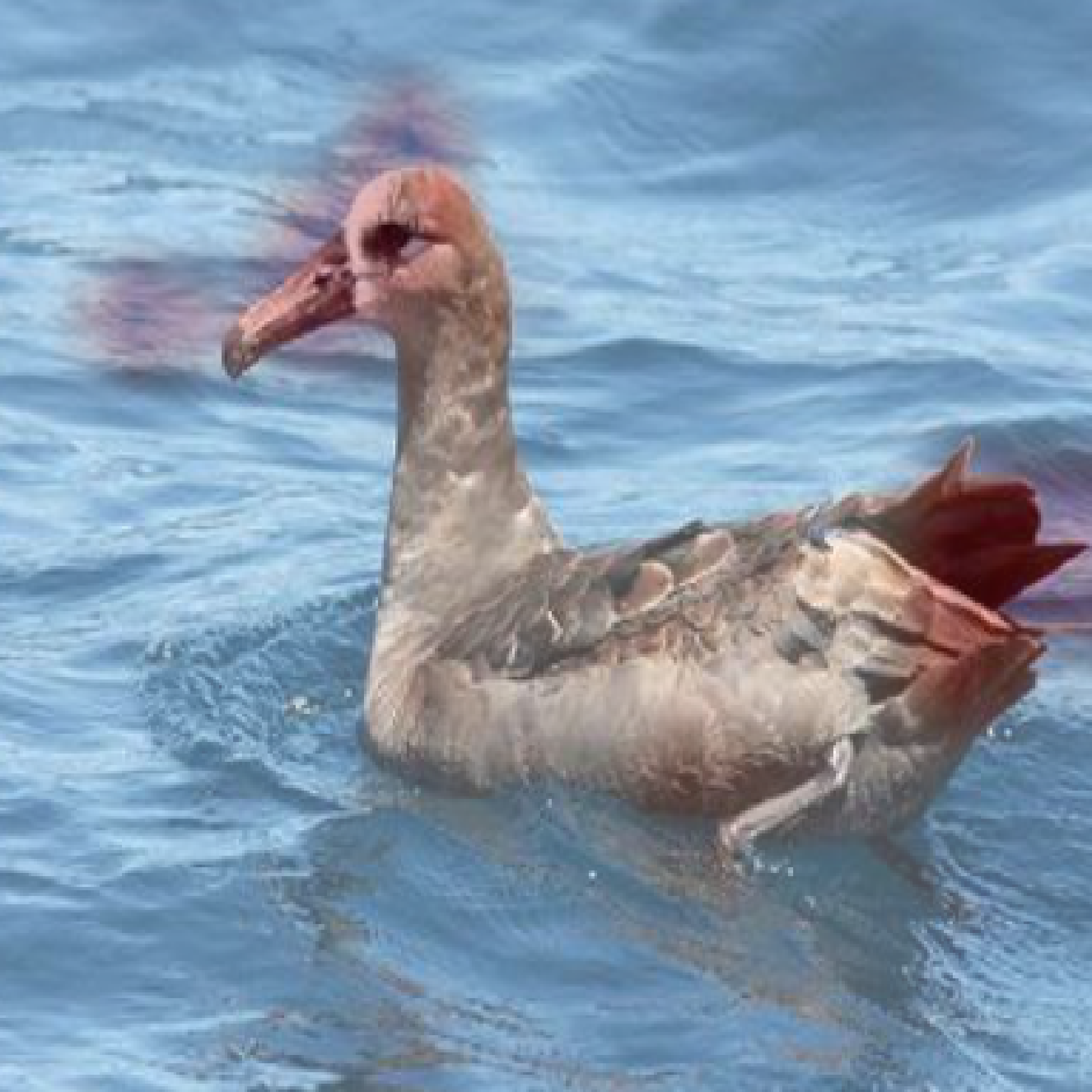}
    \end{subfigure}
    \hfill
    \begin{subfigure}[b]{0.19\linewidth}
        \centering
        \includegraphics[width=\linewidth]{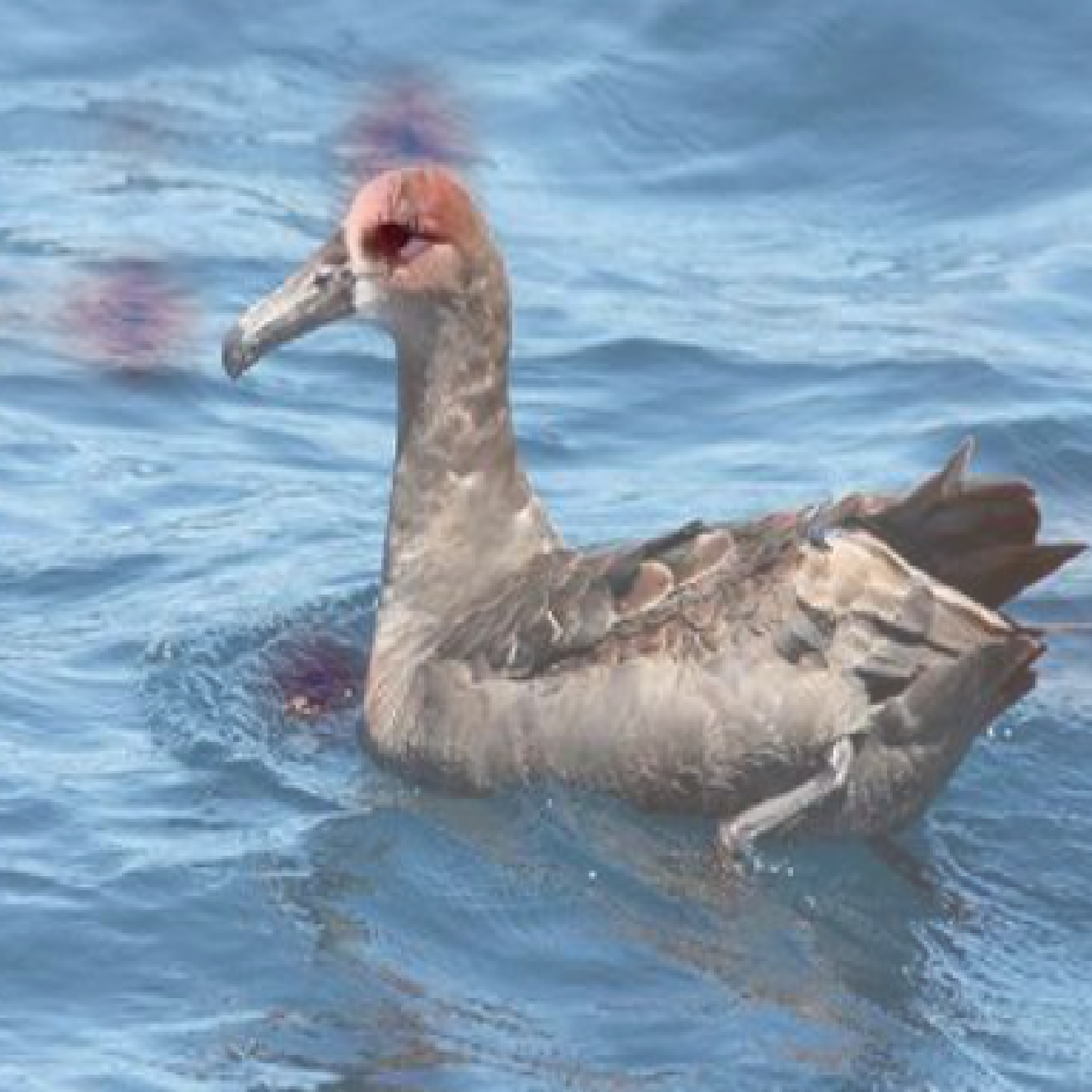}
    \end{subfigure}
    \hfill
    \begin{subfigure}[b]{0.19\linewidth}
        \centering
        \includegraphics[width=\linewidth]{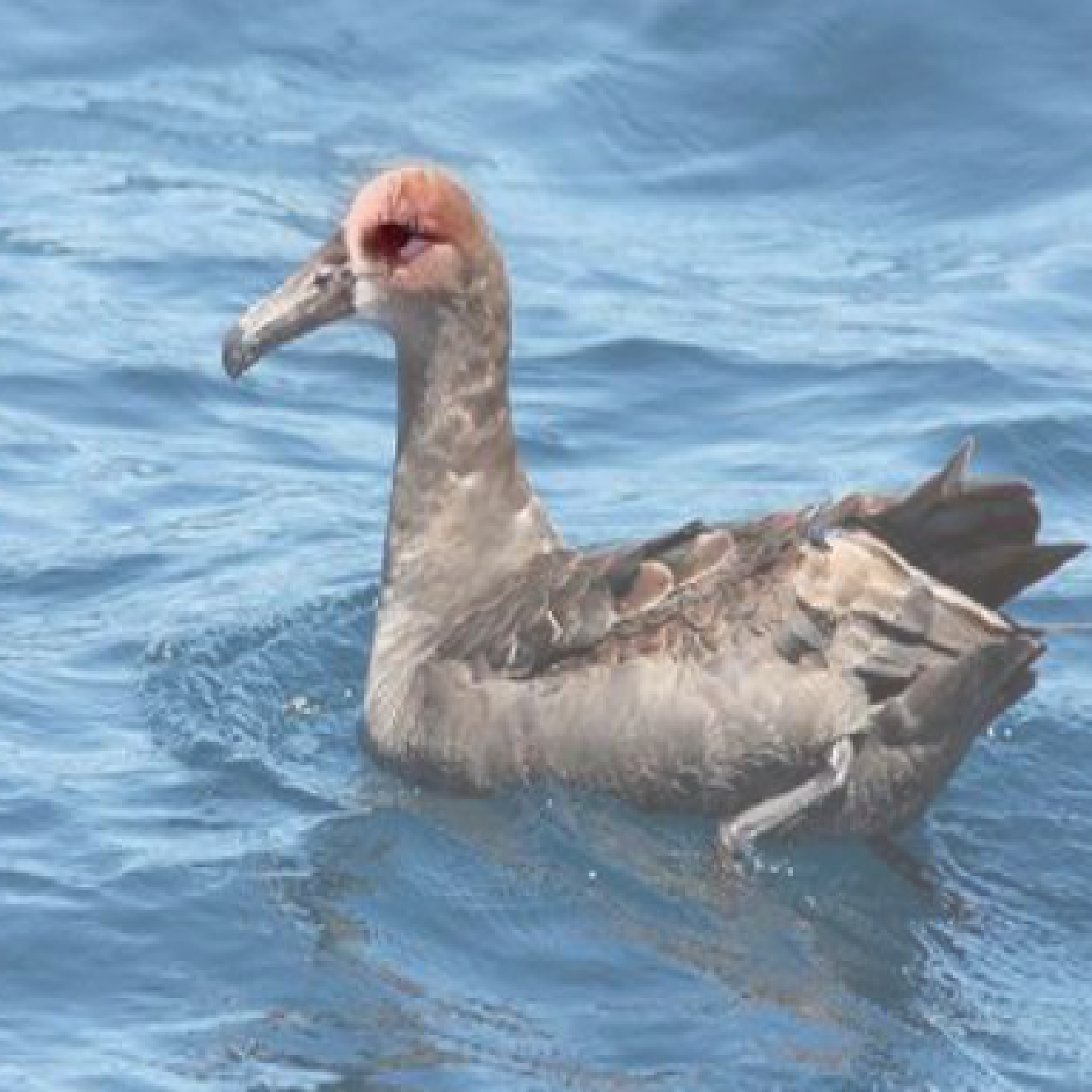}
    \end{subfigure}
    \hfill
    \begin{subfigure}[b]{0.19\linewidth}
        \centering
        \includegraphics[width=\linewidth]{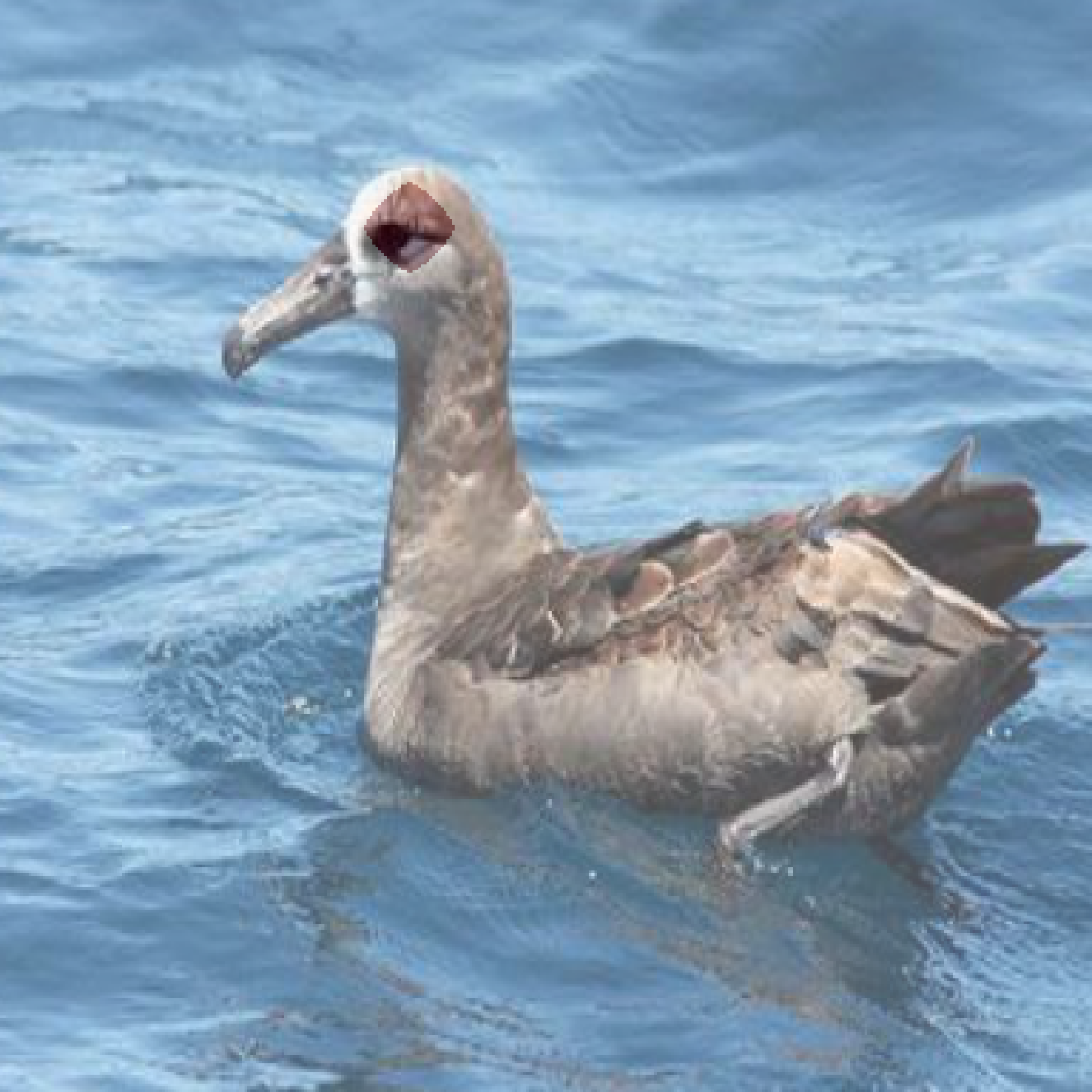}
    \end{subfigure}
    ~
    \raisebox{0.5in}{\rotatebox{90}{Brown wings}}
    \begin{subfigure}[b]{0.19\linewidth}
        \centering
        \includegraphics[width=\linewidth]{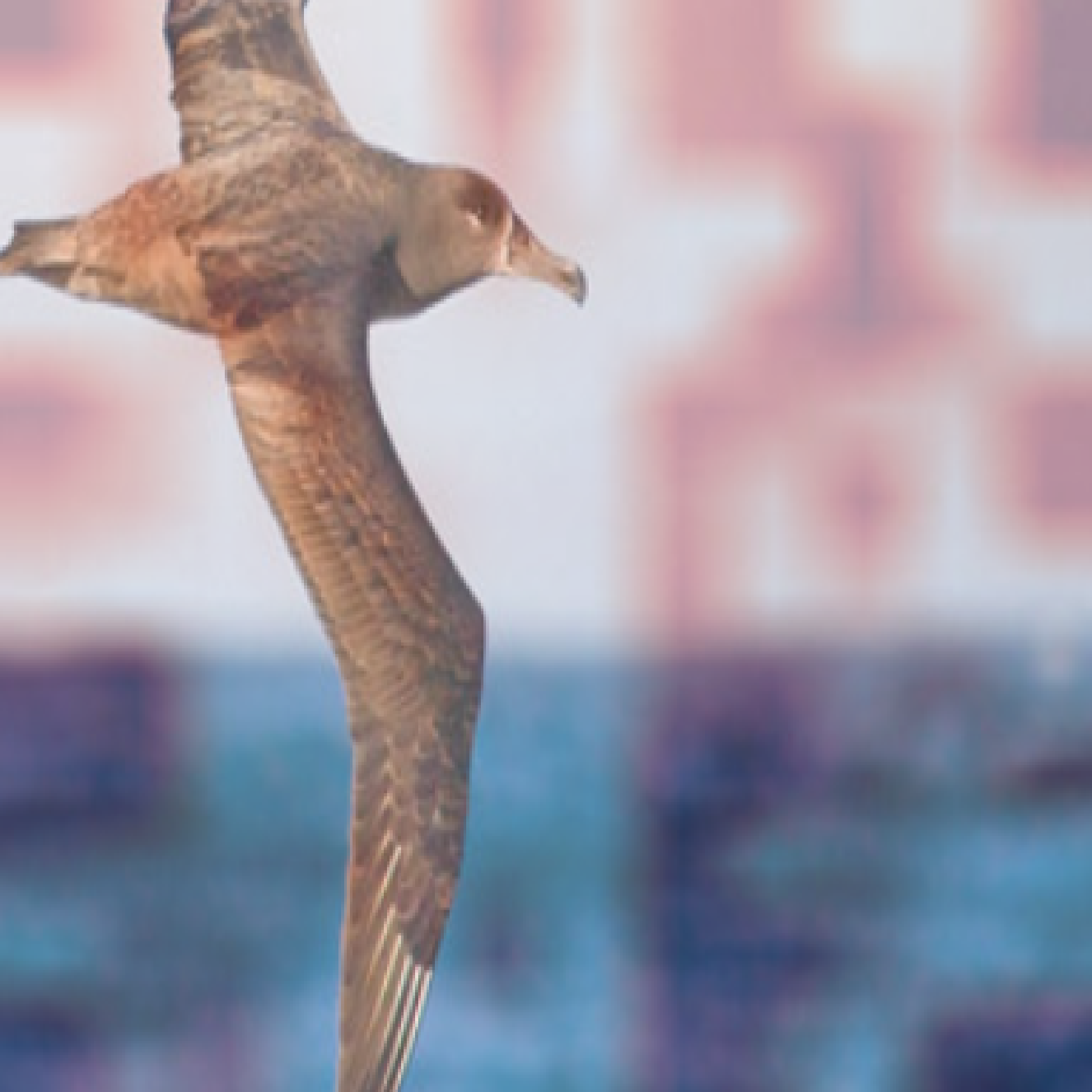}
        \caption{Zero-shot}
        \label{fig:viz-zeroshot}
    \end{subfigure}
    \hfill
    \begin{subfigure}[b]{0.19\linewidth}
        \centering
        \includegraphics[width=\linewidth]{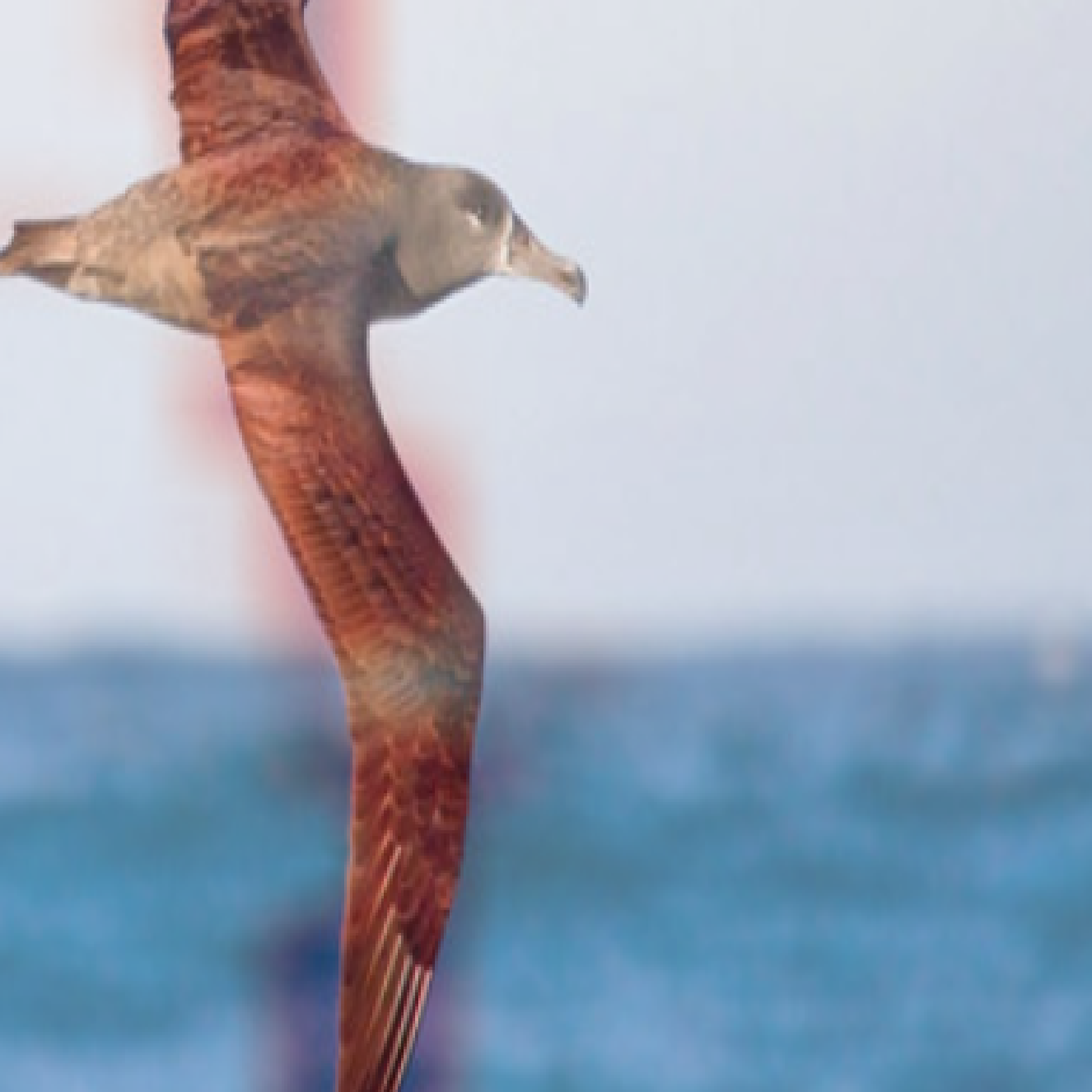}
        \caption{Unified-IO}
        \label{fig:viz-uio}
    \end{subfigure}
    \hfill
    \begin{subfigure}[b]{0.19\linewidth}
        \centering
        \includegraphics[width=\linewidth]{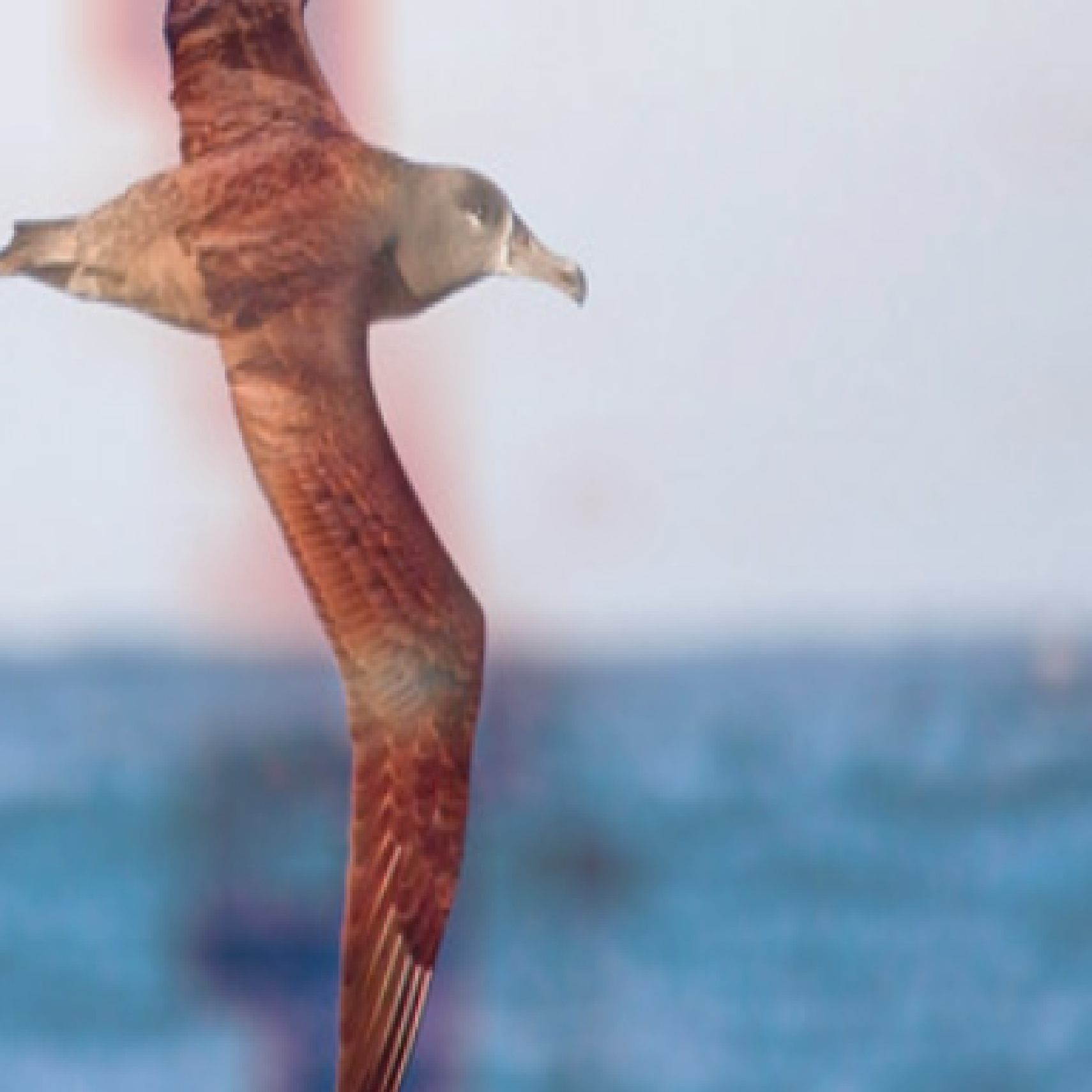}
        \caption{OFA}
        \label{fig:viz-ofa}
    \end{subfigure}
    \hfill
    \begin{subfigure}[b]{0.19\linewidth}
        \centering
        \includegraphics[width=\linewidth]{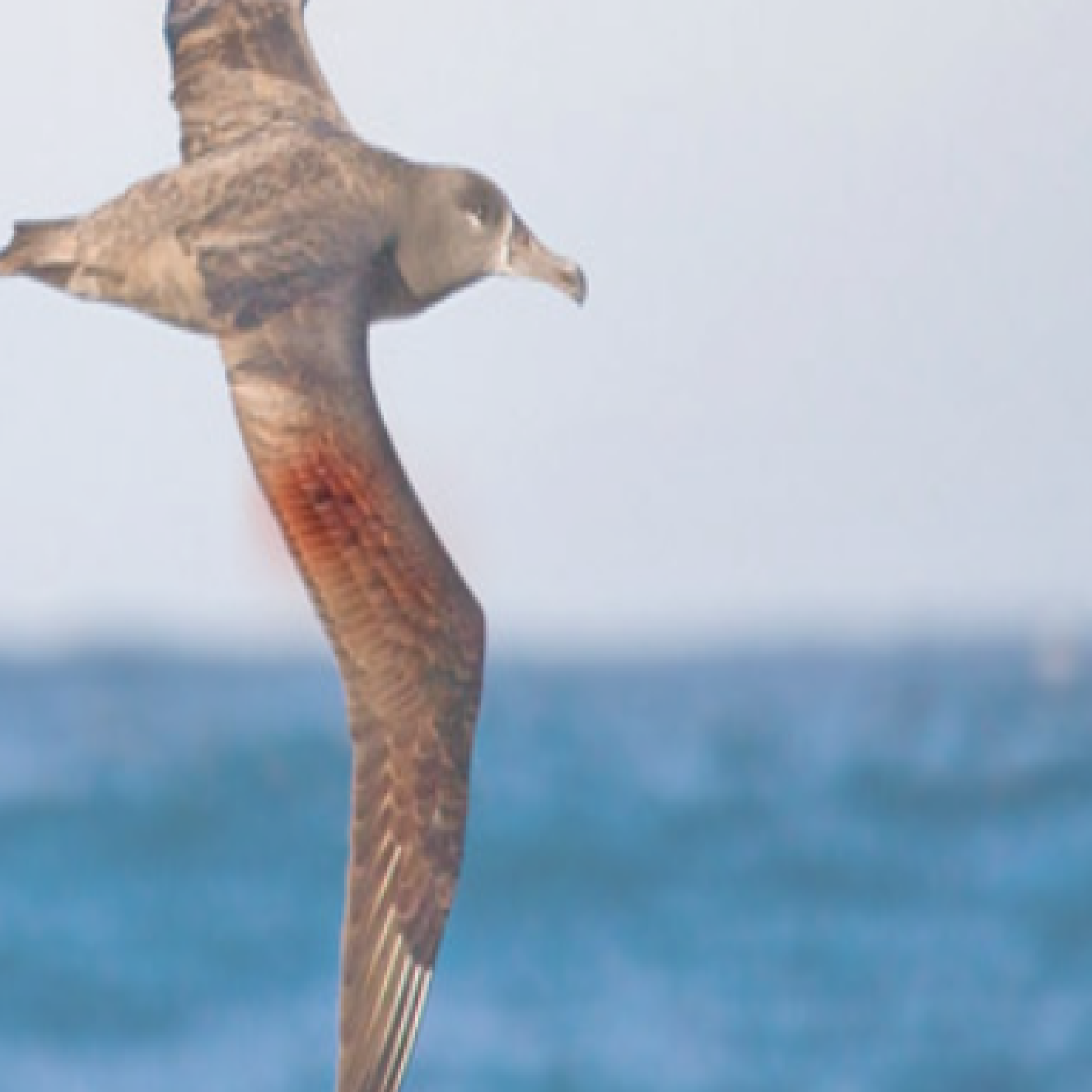}
        \caption{Expert}
        \label{fig:viz-expert}
    \end{subfigure}
    \hfill
    \begin{subfigure}[b]{0.19\linewidth}
        \centering
        \includegraphics[width=\linewidth]{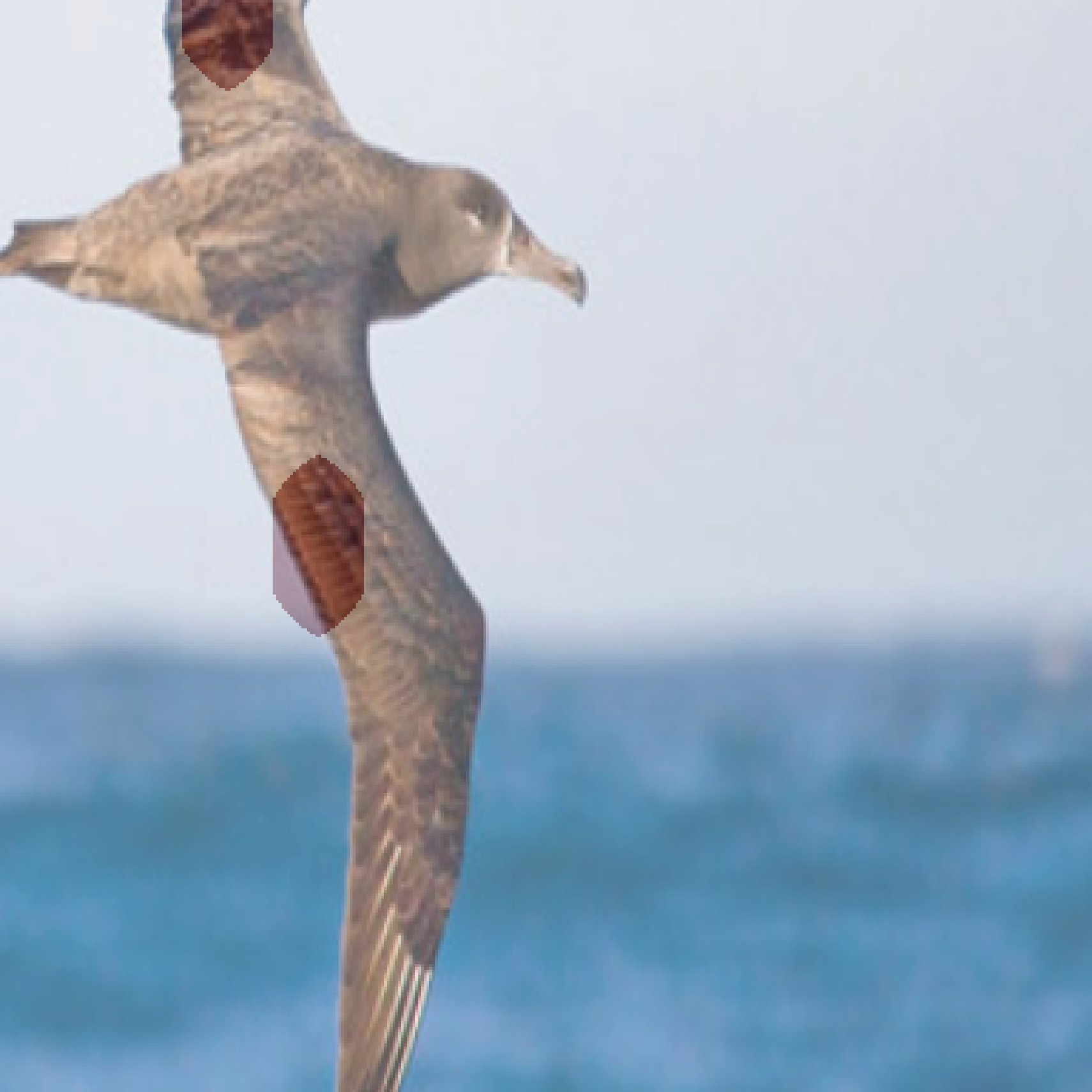}
        \caption{Ground truth}
        \label{fig:viz-annotation}
    \end{subfigure}
    \caption{Impact of localization annotation. Zero-shot predictions are shown in \ref{fig:viz-zeroshot}, fine-tuning predictions are shown in \ref{fig:viz-uio}, \ref{fig:viz-ofa} and \ref{fig:viz-expert} for localization labels from Unified-IO, OFA and experts, respectively. Expert annotations are displayed in \ref{fig:viz-annotation}.}
    \label{fig:loc-annotation}
\end{figure*}

\subsubsection*{Re-aligning text embeddings with state-of-the-art image encoder}

State-of-the-art classification results on the CUB200 dataset are obtained with a Swin-L backbone \cite{chou2022,chou2023}, pre-trained on ImageNet-21K (144 M training samples) \cite{ridnik2021}. This let us to think that a Swin-L backbone could be more suited for the given task. In the last row of Table \ref{tab:zeroshot}, we replace the ViT-L backbone of CLIP (pre-trained on LAION-2B) by a Swin-L backbone, before fine-tuning. This follows the idea of merging two independent pre-trained backbones by fine-tuning them altogether, as explored in \cite{alayrac2022}.
The whole image backbone is been fine-tuned, in addition to the last layers of the text encoder. Indeed, we did not encounter training instability with this backbone. This may be due to the inductive bias of the window attention of the Swin architecture. The choice of parameters to train is discussed in Section \ref{section:ablation-study}.
As expected, zero-shot performance is very poor since the Swin-L backbone and the CLIP text encoder have not been trained altogether so far, so there is no alignment between image and text embeddings. However, fine-tuning leads to an improvement of 4.88 points of top-1 accuracy and 13.12 points of localization mAP, compared to the ViT-L counterpart. This result is interesting given that the Swin-L backbone was pre-trained with 16 times fewer examples. We use this configuration with Swin-L backbone in the remaining of the paper.

\subsection{Relaxing annotation burden}

The quantity of annotated data is a key factor in deep neural network training. However, annotations are costly to produce. In this experiment, we show that using a multimodal foundation model as oracle, \ie, to generate pseudo-labels through zero-shot predictions, can help to replace localization expert annotations for free. 

\begin{table}[ht!]
    \caption{Comparison of different foundation models used as oracle for localization on CUB200. "None" means no supervision for the localization part ($\mathcal{L}_\text{loc}$ is discarded). Metrics are expressed in percentages. }
    \centering
    \resizebox{\linewidth}{!}{
    \begin{tabular}{ c c c c c}
        \hline
        Localization  & Top-1   & Attribute  & Attribute \\
        annotations & accuracy  & detection mAP & localization mAP\\
        \hline
        None & 91.22 & 64.30 & 6.61\\
        Expert & 91.20 & 63.78 & 74.85\\
        Unified-IO$_\text{XL}$ & 91.18 & 63.69 & 47.63\\
        OFA$_\text{Huge}$ & 91.11 & 63.71 & 60.21\\
        \hline
    \end{tabular}
    }
    \label{tab:oracle}
\end{table}

\begin{table*}[ht!]
    \caption{Ablation study over the different losses. The first line corresponds to zero-shot evaluation.}
    \centering
    \resizebox{0.75\linewidth}{!}{
    \begin{tabular}{c c c c c c c c}
        \hline
        \multirow{2}{*}{$\mathcal{L}_\text{class}$} & \multirow{2}{*}{$\mathcal{L}_\text{attr}$} & \multirow{2}{*}{$\mathcal{L}_\text{loc}$} & \multirow{2}{*}{$\mathcal{L}_\text{proj}$} & \multirow{2}{*}{$\mathcal{L}_\alpha$} & Top-1   & Attribute  & Attribute \\
        & & & & &  accuracy (\%) &  detection mAP (\%) & localization mAP (\%)\\
        \hline
        \xmark & \xmark & \xmark & \xmark  & \xmark & 0.17 & 15.34 & 5.81 \\
        \cmark & \xmark & \xmark & \xmark  & \xmark &  89.97 & 22.64 & 6.29 \\
        \cmark & \cmark & \xmark & \xmark  & \xmark &  90.02 & 65.93 & 6.81 \\
        \cmark & \cmark & \cmark & \xmark  & \xmark & 89.83 & 65.04 & 77.32\\
        \cmark & \cmark & \cmark & \cmark  & \xmark & 91.20 & 63.78 & 74.85\\
        \cmark & \cmark & \cmark & \cmark  & \cmark & 91.09 & 63.62 & 74.43\\
        \hline
    \end{tabular}
    }
    \label{tab:ablation-study}
\end{table*}

To this end, we used the largest version of two MMFM which enable free-text prompt as input: Unified-IO$_\text{XL}$ and OFA$_\text{Huge}$, which include 2925M and 930M parameters, respectively. While Unified-IO generates pixel-level segmentation maps, OFA outputs bounding boxes, which are translated into segmentation maps. We convert them to patch-level binary masks for each attribute, as in the previous experiments. We follow the given standard example prompts for both models. The only prompt fine-tuning strategy we use is the addition of the word "bird" in the OFA prompt only, as it did not improve the results for Unify-IO. The prompts are as follows:\\
{\small
Unify-IO: \textit{What is the segmentation of " BODYPART " ?} \\
OFA: \textit{which region does the text " bird's BODYPART " describe?}}

Results are shown in Table \ref{tab:oracle}. The localization is always evaluated using the expert annotations, no matter the annotation used for training. OFA provides better localization results than Unified-IO in this experiment, with fewer parameters and GPU usage. As one can notice, using MMFM as oracle can be a good option to consider for fine-grained attribute localization when no annotation is available. Even if the performance is not on par with training with expert annotation, the final results are quite impressive.

Qualitative results are given in Figure \ref{fig:loc-annotation}. We used bilinear interpolation to display matching probabilities between image patches and attribute prediction/annotation over the original images. As stated previously, the zero-shot model is not able to localize attributes. Using free pseudo-labels from MMFM enables to reach interesting localization results, sometimes better than those trained with expert annotations. Indeed, since expert annotations only consists in 2D points, the wings of the birds are poorly segmented, as illustrated in the second example (Figure \ref{fig:viz-annotation}). Localization maps from MMFM are closer to what is actually expected, focusing all over the wings.

\subsection{Model analysis}

\subsubsection*{Impact of positive/negative attribute prompts}

\begin{figure}[ht]
    \centering
    \includegraphics[width=\linewidth]{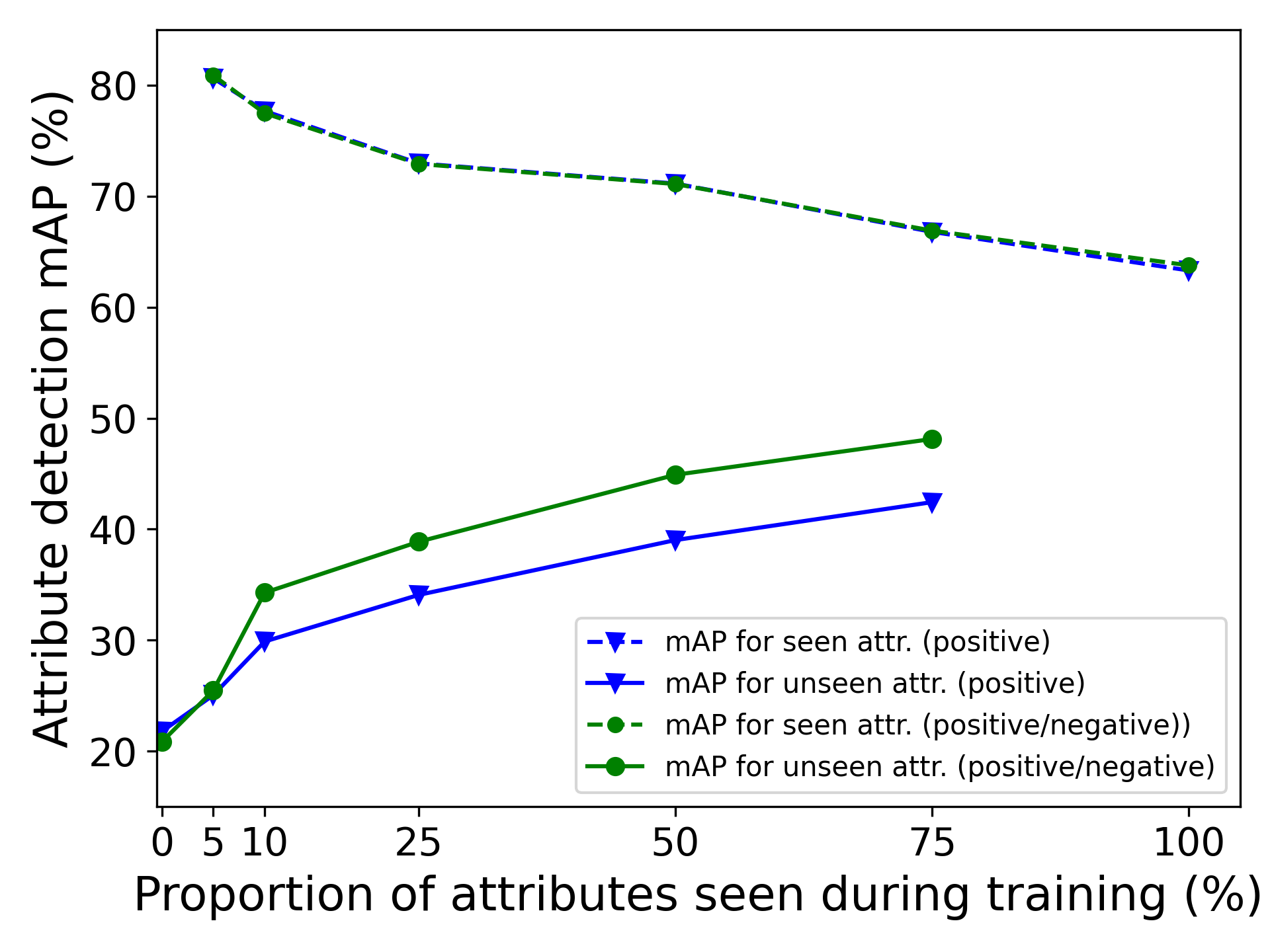}
    \caption{Impact of positive/negative prompt formulation on the detection of attributes for different ratios of seen attributes during training.}
    \label{fig:pos-neg}
\end{figure}

\begin{figure*}[ht!]
    \centering
    \begin{subfigure}[b]{0.4\linewidth}
        \centering
        \includegraphics[width=\linewidth]{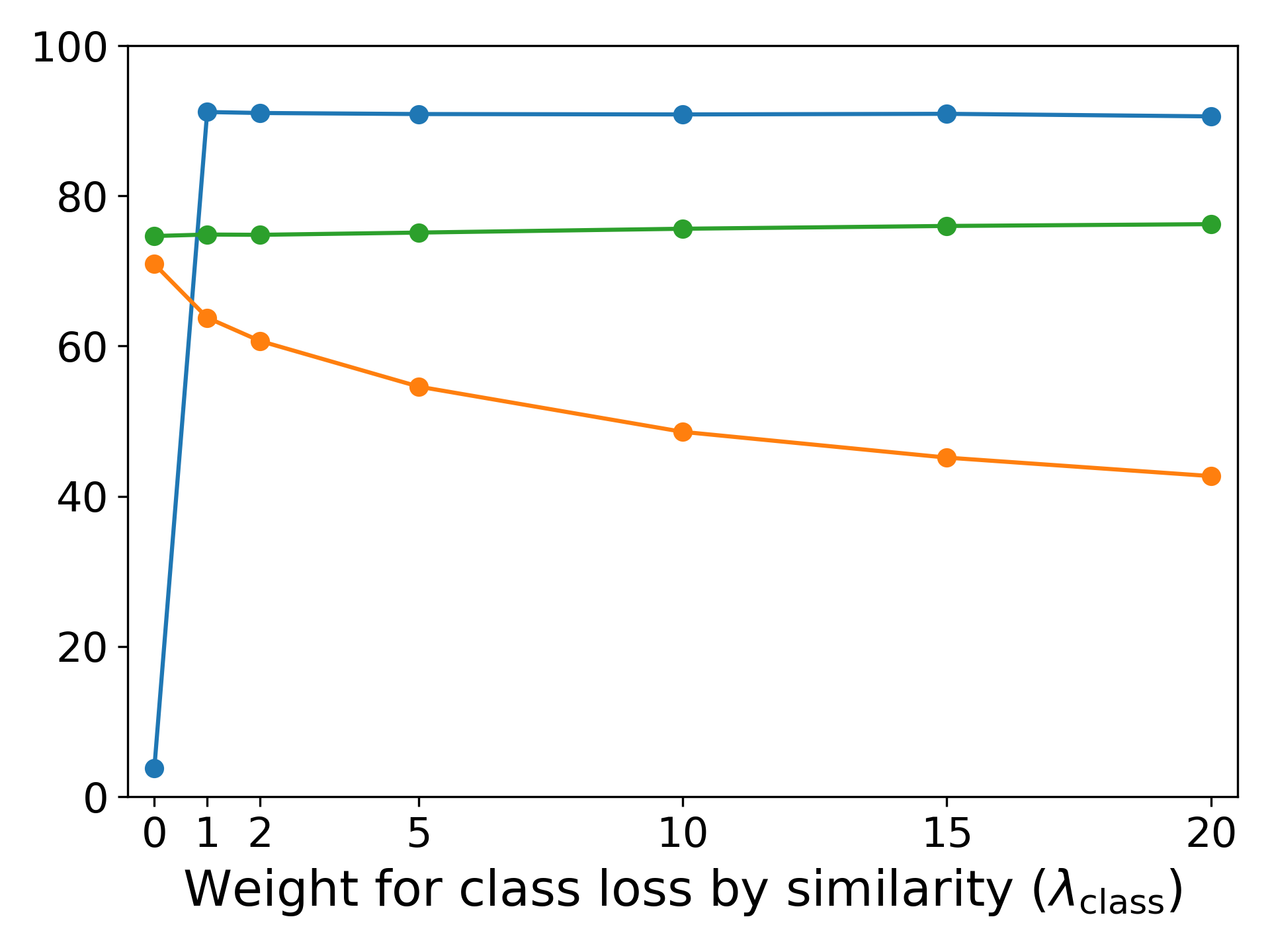}
    \end{subfigure}
    \hspace{2cm}
    \begin{subfigure}[b]{0.4\linewidth}
        \centering
        \includegraphics[width=\linewidth]{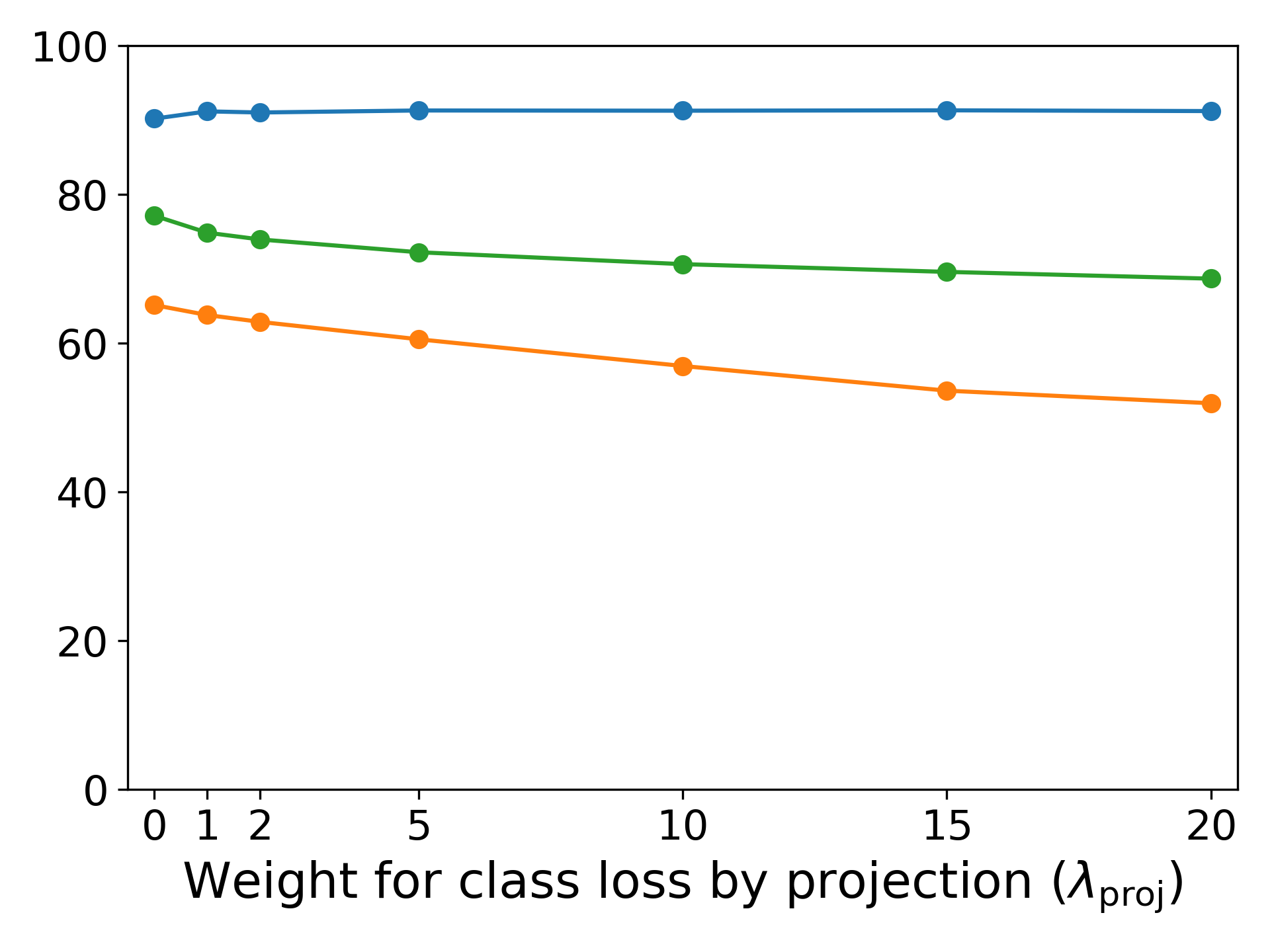}
    \end{subfigure}
    ~
    \begin{subfigure}[b]{0.4\linewidth}
        \centering
        \includegraphics[width=\linewidth]{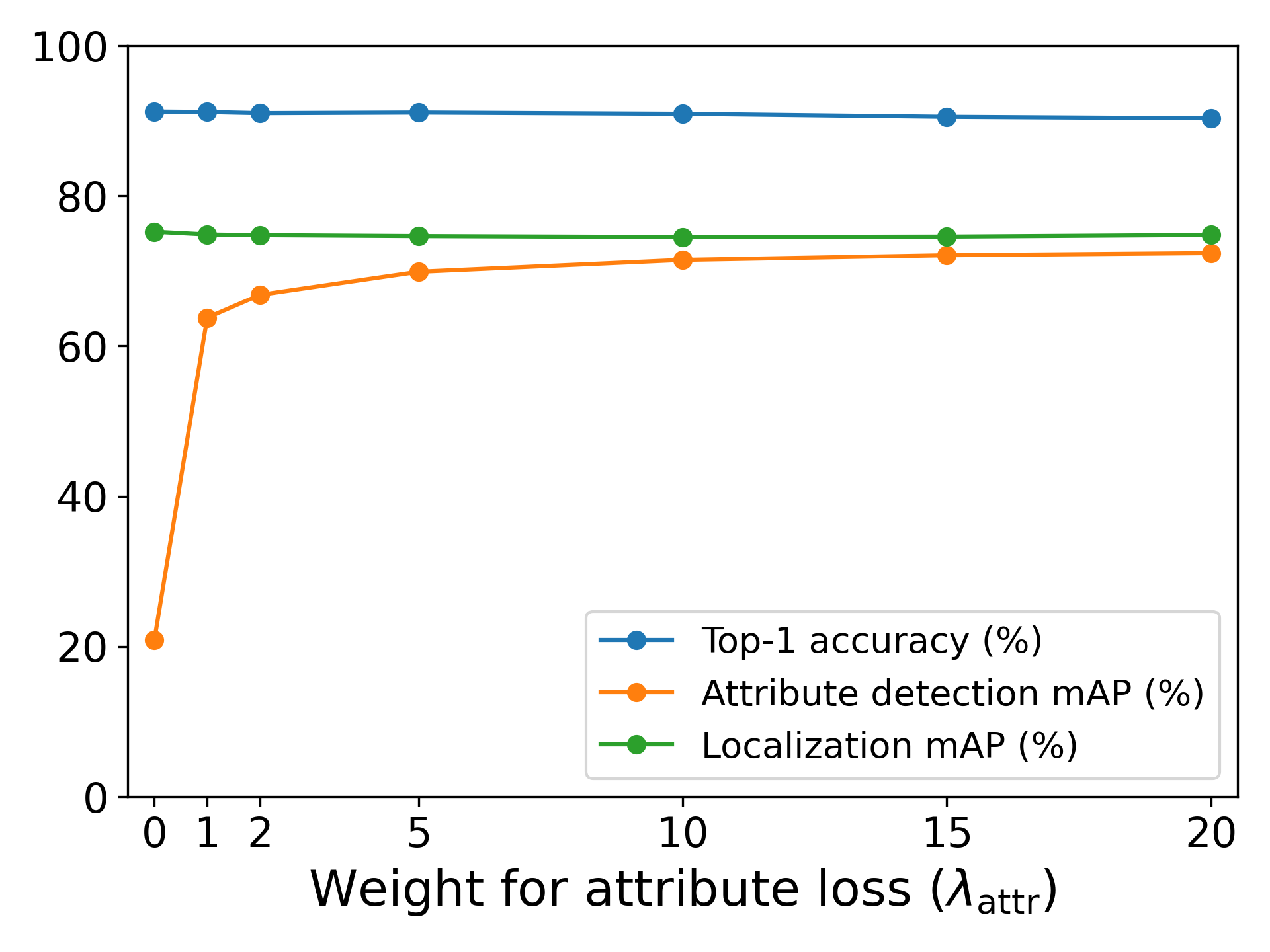}
    \end{subfigure}
    \hspace{2cm}
    \begin{subfigure}[b]{0.4\linewidth}
        \centering
        \includegraphics[width=\linewidth]{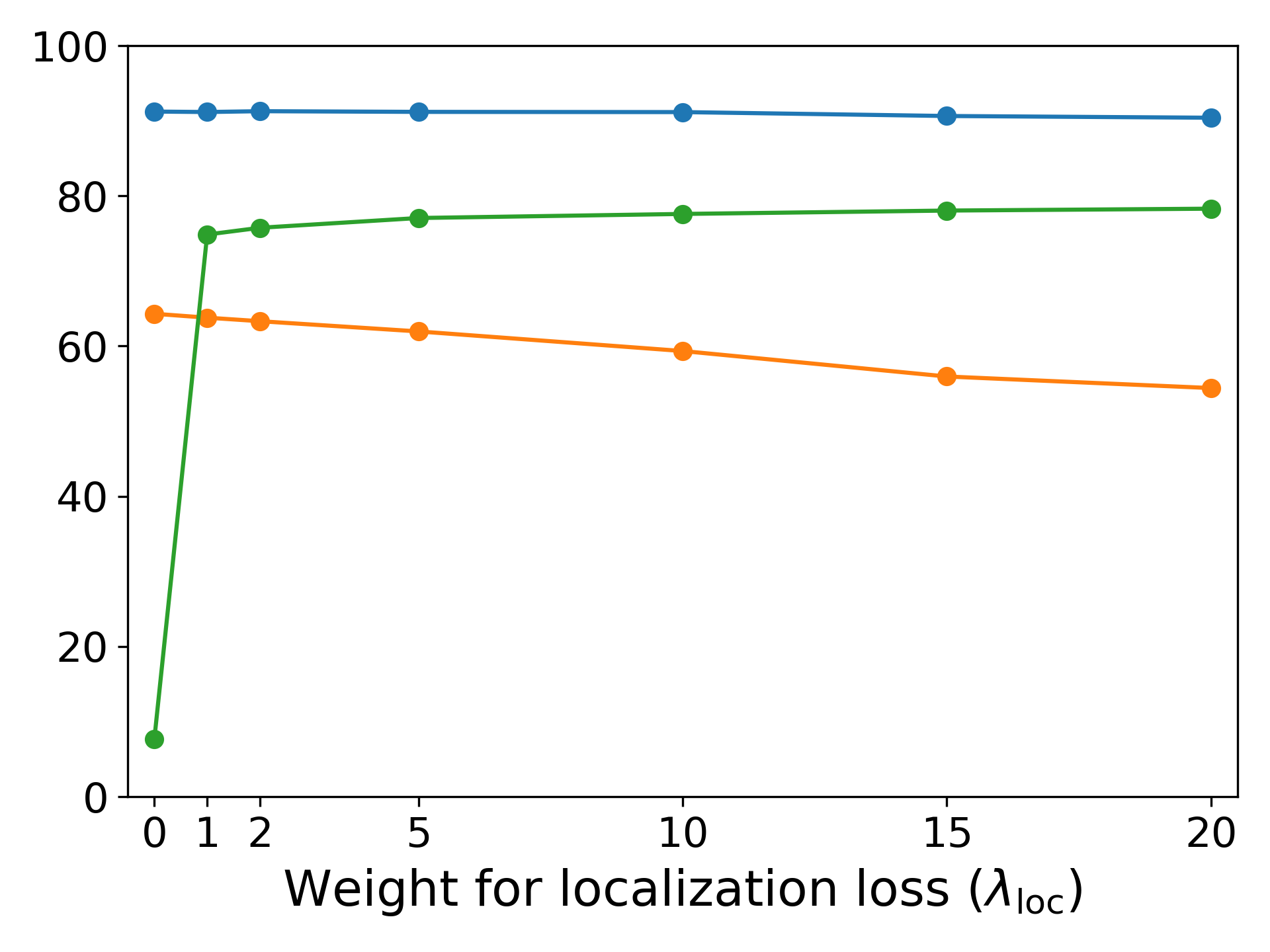}
    \end{subfigure}
    \caption{Impact of loss weighting. For each graph, all loss weights are set to unit except one which varies from 0 to 20.}
    \label{fig:loss-weights}
\end{figure*}

\begin{table*}[ht!]
    \caption{Comparison of fine-tuning strategies. In any cases, the last transformer and projection layers are trained, for both image and text encoders. This experiment focuses on the usefulness of training the whole backbones. Metrics are in percentages.}
    \centering
    \resizebox{0.75\linewidth}{!}{
    \begin{tabular}{c c c c c c c}
        \hline
        Training image & Training text & Top-1 & Attribute  & Attribute & \# trained \\
        backbone & backbone &  accuracy & detection mAP & localization mAP & params\\
        \hline
        \xmark & \xmark & 88.64 & 63.65 & 43.33 & 39 M (12\%)\\
        \xmark & \cmark & 87.78 & 70.67 & 37.59 & 155 M (48\%)\\
        \cmark & \xmark & 91.20 & 63.78 & 74.85 & 205 M (64\%)\\
        \cmark & \cmark & 89.97 & 71.66 & 73.93 & 321 M (100\%)\\
        \hline
    \end{tabular}
    }
    \label{tab:fine-tuning}
\end{table*}

In this experiment, we study the impact of the combination of positive and negative formulation for attribute prompts. Initially used in zero-shot context \cite{pellegrini2023}, we show that this strategy can also be beneficial when fine-tuning, especially for the detection of unseen attributes during training. We train the model with different ratios of seen/unseen attributes, with two different configurations: with positive/negative formulation (as described in Section \ref{section:method-attributes}), or with positive prompt only. It means that for all image samples, the model is only trained with a subset of the 312 original attributes. Results are illustrated in Figure \ref{fig:pos-neg}. We compute the attribute detection mAP for both seen and unseen attributes. As one can note, the positive/negative formulation increases the attribute detection for unseen attributes of at least 4 points as of 10\% of seen attributes, while preserving the performance on seen attributes.

\subsubsection*{Impact of the hybrid loss}
\label{section:ablation-study}

In Table \ref{tab:ablation-study}, we provide an ablation study over the different losses which are combined. In this experiment, all losses are either discarded (\xmark), or weighted with unit factor (\cmark). We also compare the model in zero-shot (no loss), and, as expected, the results are very poor for each metric. 
As one can note, successively adding the main losses ($\mathcal{L}_\text{class}$, $\mathcal{L}_\text{attr}$ and $\mathcal{L}_\text{loc}$) enables to learn the three target tasks: classification, attribute detection and attribute localization, respectively. 

Adding a second loss $\mathcal{L}_\text{proj}$ for the classification task, by projecting the image embedding, leads to an improvement of 1.33 points of top-1 accuracy, while keeping competitive performance on the other two tasks. It has to be noted that the two additional projection layers (for this loss and for localization) only represent a 0.4\% increase in the number of parameters.
As illustrated by the last row of the table, adding a second loss $\mathcal{L}_\alpha$ for the attribute detection task, in a similar way as for the classification task, does not lead to any improvement, so we discarded it.

\subsubsection*{Impact of loss weighting}
We study the choice of weights for the losses. For each loss, we train the model with different values for its weight (0, 1, 2, 5, 10, 15 and 20), while keeping the weight of the other three losses to unit. Results are shown in Figure \ref{fig:loss-weights}. As one can note, classification and localization performance are rather robust to the variation of loss weights. Unit weights seem to be a good choice, except for the attribute detection loss. Setting $\mathcal{L}_\text{attr}$ to 10 leads to 91.09\% of top-1 accuracy, 69.89\% of attribute detection mAP and 74.64\% of localization mAP.

\subsubsection*{Impact of fine-tuning strategy}
We now discuss the fine-tuning strategy in terms of trainable weights. To this end, image and text backbones are alternatively either completely fine-tuned (\cmark), or only fine-tuned as of the last transformer layer (\xmark). 
Results are shown in Table \ref{tab:fine-tuning}. As one can note, fine-tuning the whole image backbone is necessary to reach interesting performance for the localization task, leading to an increase of more than 30 points. Only fine-tuning the last layer of the text encoder enables to reach better classification (91.16\% compared to 89.97\%) and localization results, to the detriment of attribute location (-7.88 points of mAP).

\section{Conclusion}
In this paper, we propose a new multitask fine-tuning strategy to enhance fine-grained classification through the detection and localization of attributes. Dedicated to multimodal foundation models, we demonstrate the effectiveness of this approach with the CLIP architecture on the CUB200 dataset. We show different ways of leveraging foundation models: through fine-tuning, by combining backbones from two different pre-training (Swin and CLIP), and through zero-shot by generating pseudo-labels and alleviating the need for annotations. Finally, we prove that the positive/negative prompt formulation is beneficial for the detection of unseen concepts during training.

\subsubsection*{Acknowledgments}
This work was granted access to the HPC resources of IDRIS under the allocation 2022-AD011013888.

{\small
\bibliographystyle{ieee_fullname}
\bibliography{References.bib}
}

\end{document}